\documentclass[10pt,twocolumn,letterpaper]{article}

\usepackage[pagenumbers]{cvpr} %

\usepackage{amsmath,amsfonts,amsthm,amssymb}
\usepackage{mathtools}
\usepackage{bm}
\usepackage{nicefrac}
\usepackage{microtype}
\usepackage{lipsum}

\usepackage{color,xcolor}
\usepackage{epsfig}
\usepackage{graphicx}

\usepackage{adjustbox}
\usepackage{array}
\usepackage{booktabs}
\usepackage{colortbl}
\usepackage{wrapfig}
\usepackage{hhline}
\usepackage{multirow}
\usepackage{subcaption}
\usepackage[size=small]{caption}
\usepackage{float}

\usepackage{changepage}
\usepackage{extramarks}
\usepackage{fancyhdr}
\usepackage{lastpage}
\usepackage{setspace}
\usepackage{soul}
\usepackage{xspace}

\usepackage{url}
\usepackage{algorithm}
\usepackage{algorithmicx}
\usepackage{algpseudocode}
\usepackage{enumerate}
\usepackage{enumitem}  %
\usepackage{makecell}
\usepackage{pifont}
\usepackage{titlecaps}
\usepackage[accsupp]{axessibility}
\usepackage{framed}

\newcommand{\myparagraph}[1]{%
    \vspace{0.1cm}
    \noindent
    \textbf{#1}%
}
\newcommand{\myfirstparagraph}[1]{%
    \noindent
    \textbf{#1}%
}

\definecolor{cvprblue}{rgb}{0.21,0.49,0.74}
\definecolor{Cardinal}{rgb}{0.549,0.082,0.082}

\definecolor{MyDarkBlue}{rgb}{0,0.08,1}
\definecolor{MyAqua}{rgb}{0,0.7,0.7}
\definecolor{MyDarkGreen}{rgb}{0.02,0.6,0.02}
\definecolor{MyDarkRed}{rgb}{0.8,0.02,0.02}
\definecolor{MyDarkOrange}{rgb}{0.40,0.2,0.02}
\definecolor{MyPurple}{RGB}{111,0,255}
\definecolor{MyRed}{rgb}{1.0,0.0,0.0}
\definecolor{MyGold}{rgb}{0.75,0.6,0.12}
\definecolor{MyDarkgray}{rgb}{0.66, 0.66, 0.66}
\definecolor{MyGreen}{rgb}{0.00, 1.00, 0.68}

\newcommand{\model}{AnyLift\xspace}

\definecolor{cvprblue}{rgb}{0.21,0.49,0.74}
\usepackage[pagebackref,breaklinks,colorlinks,allcolors=cvprblue]{hyperref}

\usepackage[normalem]{ulem}

\def\paperID{8315} %
\def\confName{CVPR}
\def\confYear{2026}

\title{AnyLift: Scaling Motion Reconstruction from Internet Videos via 2D Diffusion\vspace{-8pt}} 

\author{
    Hongjie Li$^{*,\dagger}$ \qquad 
    Heng Yu$^{*}$ \qquad 
    Jiaman Li \qquad 
    Hong-Xing Yu \\
    Ehsan Adeli \qquad
    C. Karen Liu \qquad
    Jiajun Wu \\[0.5em]
    Stanford University
    \vspace{-21pt}
}

\begin{document}

\twocolumn[{%
\renewcommand\twocolumn[1][]{#1}%
\maketitle
\begin{center}
    \centering
    \vspace{2pt}
    \captionsetup{type=figure}
    \includegraphics[width=1\textwidth]{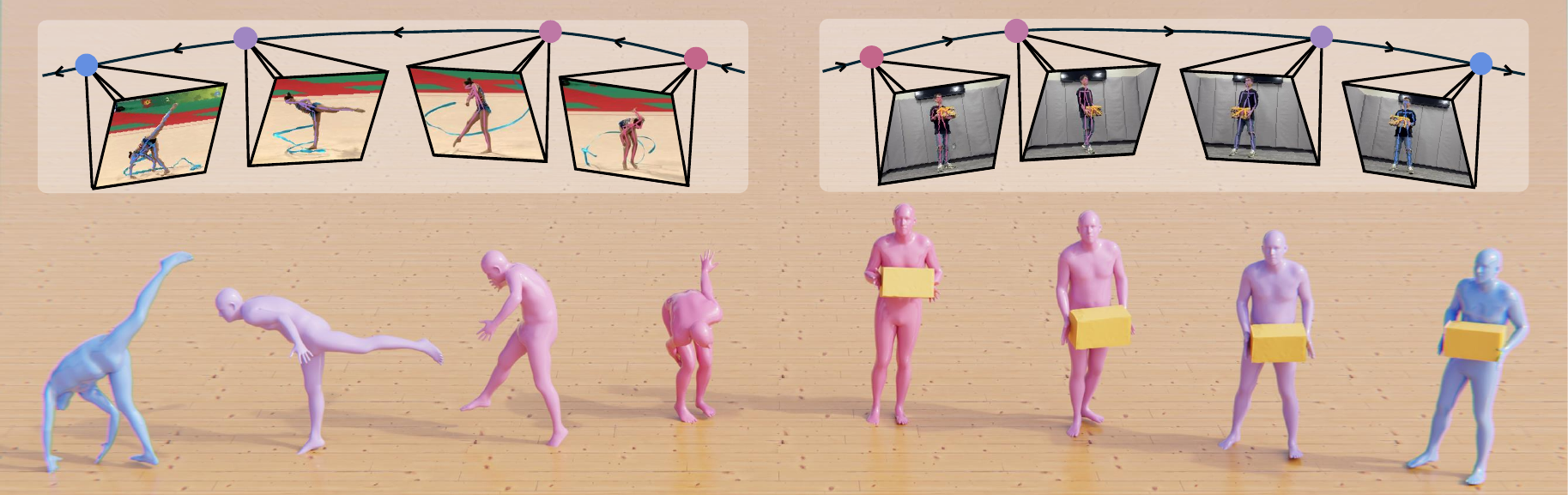}
    \vspace{-0.6cm}
    \caption{\textbf{Human and human-object interaction (HOI) motions lifted by our approach.} Trained on 2D keypoints and corresponding camera trajectories, our framework \model reconstructs world‑coordinated 3D human motion and HOI from monocular videos captured by dynamic cameras. We demonstrate its effectiveness on human motion reconstruction from Internet gymnastics videos (left) and on HOI reconstruction from captured real-world videos (right).  Please refer to our \href{https://awfuact.github.io/anylift/}{\textit{project page}} for video results.
    }
    \label{fig:teaser}
\end{center}%
}]

{
\let\thefootnote\relax\footnotetext{
$^*$ Equal contribution. 
$^\dagger$ Work was done while H. Li was a visiting student at Stanford University. 
H. Li is now with Peking University.
}
}
\begin{abstract}
Reconstructing 3D human motion and human-object interactions (HOI) from Internet videos is a fundamental step toward building large-scale datasets of human behavior. Existing methods struggle to recover globally consistent 3D motion under dynamic cameras, especially for motion types underrepresented in current motion-capture datasets, and face additional difficulty recovering coherent human-object interactions in 3D. We introduce a two-stage framework leveraging 2D diffusion that reconstructs 3D human motion and HOI from Internet videos. In the first stage, we synthesize multi-view 2D motion data for each domain, leveraging 2D keypoints extracted from Internet videos to incorporate human motions that rarely appear in existing MoCap datasets. In the second stage, a camera-conditioned multi-view 2D motion diffusion model is trained on the domain-specific synthetic data to recover 3D human motion and 3D HOI in the world space. We demonstrate the effectiveness of our method on Internet videos featuring challenging motions such as gymnastics, as well as in-the-wild HOI videos, and show that it outperforms prior work in producing realistic human motion and human-object interaction. \vspace{-12pt}
\end{abstract}

\section{Introduction}

Large-scale 3D human motion and human-object interaction (HOI) data are essential for a wide range of applications in computer vision, computer graphics, and robotics. Such data enable realistic character animation, simulation of human behavior in virtual environments, and motion imitation or policy learning for humanoid robots. While high-quality motion capture (MoCap) datasets have been widely used for these purposes, they remain limited in scale and diversity. Acquiring MoCap data requires controlled environments, specialized hardware, and professional actors, making it infeasible to capture the vast diversity of motions observed in everyday life. Estimating 3D human motion directly from videos offers a scalable alternative. However, while recent progress has been made in lifting 2D pose sequences to 3D motion, existing methods still struggle to reconstruct global motion in the world coordinate frame under dynamic camera settings, especially for motion categories that are rarely represented in existing MoCap datasets, as well as human-object interactions involving dynamic object movement. 

Existing approaches to human motion estimation can be broadly grouped into two categories.
The first relies on large-scale MoCap datasets~\cite{mahmood2019amass} to train networks with 3D supervision~\cite{shin2024wham,shen2024world}.
Although these models achieve high accuracy for in-distribution motion, they generalize poorly to actions that are underrepresented in MoCap data, such as gymnastics, martial arts, or other dynamic, real-world movements, since collecting such 3D data is prohibitively difficult.
The second direction seeks to reduce dependence on 3D data by leveraging 2D keypoints extracted from domain-specific videos~\cite{kapon2024mas,li2025lifting}. 
For example, MVLift~\cite{li2025lifting} reconstructs 3D motion using only 2D inputs within a multi-stage generative framework.
While effective under static cameras, this approach assumes fixed viewpoints during both training and inference, whereas real-world videos commonly feature moving cameras and limited view coverage.
These constraints make it difficult to scale to diverse in-the-wild videos where camera motion varies significantly across sequences.
Furthermore, extending these frameworks to reconstruct world-grounded human-object interaction from in-the-wild videos remains an open problem.

In this work, we adopt the formulation of learning 2D motion priors for 3D reconstruction~\cite{li2025lifting}, as it scales beyond the limited motion diversity of MoCap datasets and naturally incorporates HOI within the same reconstruction framework as human motion. Specifically, we present a unified two-stage framework that reconstructs both 3D human motion and 3D human-object interactions (HOI) from dynamic-camera videos.
In Stage 1, we synthesize multi-view 2D motions that serve as training data for the subsequent stage. In Stage 2, we train a camera-conditioned multi-view 2D diffusion model on the synthesized data to reconstruct 3D motion directly from single-view 2D keypoints. Although human motion and HOI reconstruction share the same two-stage structure, they differ in how the synthetic multi-view data are generated. For human motions that are underrepresented in existing MoCap datasets, we leverage Internet videos to generate synthetic multi-view data by learning single-view 2D motion priors and introducing a hybrid data-source training strategy that mitigates limited viewpoint coverage. For HOI, we focus on reconstruction of everyday interactions. Although existing 3D HOI MoCap datasets contain only a limited set of objects, HOI motions often exhibit consistent interaction patterns across objects within the same category. Therefore, we synthesize category-specific multi-view 2D trajectories by reprojecting existing HOI MoCap sequences under diverse camera trajectories.
Given the resulting synthetic data, our Stage-2 diffusion model predicts consistent multi-view 2D motions from a single-view input, enabling faithful reconstruction of global 3D human motion and HOI in world coordinates.

To summarize, our work makes the following contributions.
First, we propose a camera-trajectory-conditioned 2D motion diffusion model that enables the use of dynamic-camera videos for training and motion reconstruction from unconstrained monocular inputs.
Second, we introduce a hybrid data-source training strategy with a decomposed motion representation to address the challenge of limited camera-view coverage during training.
Third, we present a unified framework for reconstructing world-grounded human motion and human-object interactions from in-the-wild videos. 

\section{Related Work}

\myfirstparagraph{Human Motion Reconstruction from Video.}
Estimating 3D human motion from monocular videos has been widely studied using the SMPL~\cite{loper2015smpl} body model, with advances in both image-based~\cite{kanazawa2018end,kolotouros2019learning,kocabas2021pare,li2021hybrik,zhang2023pymaf} and video-based methods~\cite{kocabas2019vibe,wan2021encoder,shen2023global,shin2024wham}. Recent works~\cite{yuan2022glamr,ye2023decoupling,sun2023trace,kocabas2024pace,shin2024wham} focus on recovering global trajectories to obtain human motion in the world coordinate system. However, these methods are typically trained with 3D motion capture datasets such as AMASS~\cite{mahmood2019amass}, which limits their generalization to motion types underrepresented in MoCap. In contrast, our approach learns 2D motion priors directly from Internet videos, allowing reconstruction of diverse motions such as gymnastics and martial arts that are rarely captured in MoCap collections.

\myparagraph{Human-Object Interaction Reconstruction from Video.}
Reconstructing human-object interactions from videos has been less explored compared to human motion alone. Recent work has jointly modeled human motion, object motion, and contact~\cite{xie2022chore,xie2023visibility}.
CHORE~\cite{xie2022chore} estimates human and object poses from a single image.
VisTracker~\cite{xie2023visibility} reconstructs human, object, and contact trajectories from a single RGB camera by conditioning neural field representations on SMPL fits and performing visibility-aware temporal aggregation. While robust to occlusion, it assumes a static camera and focuses on relative human-object motion rather than world-coordinate trajectories under moving cameras. Our work instead reconstructs both human and object motion in the world coordinate system, enabling interaction reconstruction under dynamic camera settings.

\myparagraph{Weakly-Supervised Motion Estimation.}
To mitigate dependence on 3D motion capture data, several works train only on 2D poses or weak supervision. Traditional 2D-to-3D lifting methods~\cite{martinez2017simple,pavllo20193d,cai2019exploiting,wang2020motion,li2022mhformer,zhang2022mixste} regress 3D joint positions from 2D inputs using MLPs, temporal convolutions, or transformers. However, they remain constrained by paired 2D-3D training data and struggle to generalize beyond their capture domains.
Recent weakly supervised methods such as ElePose~\cite{wandt2022elepose} and MAS~\cite{kapon2024mas} train solely on in-domain 2D sequences, relying on reprojection or diffusion-based priors to recover plausible 3D poses. Despite removing the need for 3D supervision, these methods are restricted to local pose estimation and cannot infer global trajectories in the world frame.
MVLift~\cite{li2025lifting} extends this weakly supervised setting by learning a generative lifting model from 2D inputs to recover global 3D motion, yet it remains limited to static-camera scenarios and struggles when viewpoint coverage is limited during training.
Our framework advances this line of research by learning camera-conditioned 2D motion priors and introducing a hybrid training strategy that together enable reconstruction of world-coordinated human motion from unconstrained Internet videos.

\myparagraph{Multi-View 2D Content Generation.}
Large-scale 3D collections such as Objaverse~\cite{deitke2023objaverse} have spurred multi-view generation for objects, where diffusion models synthesize view-consistent images for 3D reconstruction~\cite{liu2023zero,shi2023mvdream,liu2023syncdreamer,long2024wonder3d}. These ideas have recently extended to multi-view video generation that jointly models spatial and temporal consistency for 4D content~\cite{kuang2024collaborative,van2024generative}. Inspired by prior works on multi-view generation, MVLift~\cite{li2025lifting} reformulates 3D motion reconstruction as multi-view 2D motion generation from single-view inputs, learning entirely from 2D keypoints without relying on 3D annotations.
We build on this formulation and extend it to handle dynamic cameras, employ hybrid training to address limited viewpoint coverage, and reconstruct human-object interactions from in-the-wild videos.

\begin{figure*}[t!]
    \centering
    \includegraphics[width=1\textwidth]{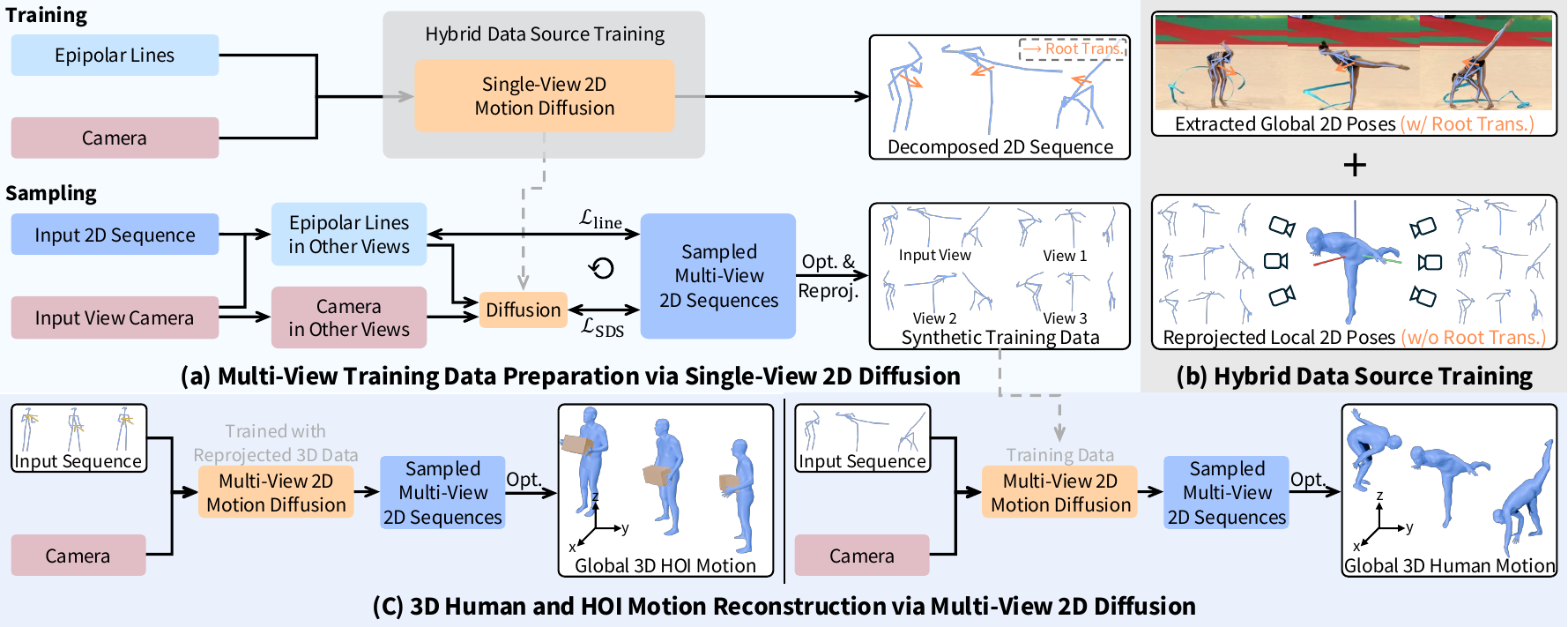}
    \vspace{-0.6cm}
    \caption{\textbf{Overview of \model.} (a) We first train a single‑view 2D motion diffusion model conditioned on camera trajectories and epipolar lines to synthesize multi‑view 2D training data. (b) During training, we employ a hybrid data source strategy that enhances viewpoint coverage by combining global 2D pose sequences from videos with locally reprojected poses. (c) Finally, we train a multi‑view 2D motion diffusion model to reconstruct consistent world‑coordinated 3D human and HOI motions from real-world videos.}
    \label{fig:method}
    \vspace{-12pt}
\end{figure*}

\section{\model}

\myfirstparagraph{Formulation.} 
Our goal is to estimate world-coordinated 3D human and human-object interaction (HOI) motion sequences $\tau = (\mathcal{H}, \mathcal{O})$ from single-view 2D keypoint sequences $\mathbf{X} \in \mathbb{R}^{T \times K \times 2}$ captured under dynamic cameras. Here, $\mathcal{H}$ denotes the human motion, and $\mathcal{O}$ denotes the object motion, which is included only for HOI reconstruction. $T$ and $K$ represent the number of frames and keypoints, respectively. We adopt the SMPL model~\cite{loper2015smpl} to parameterize the human pose, where each frame $t$ is represented by $\mathcal{H}_t = (\mathbf{r}_t, \boldsymbol{\phi}_t, \boldsymbol{\Theta}_t)$, consisting of the root translation $\mathbf{r}_t$, global orientation $\boldsymbol{\phi}_t$, and body pose parameters $\boldsymbol{\Theta}_t$.

\myparagraph{Preliminary.}
MVLift~\cite{li2025lifting} reconstructs world-coordinated 3D motion sequences from single-view 2D pose inputs. The framework enforces multi-view consistency through four stages. Stages 1-3 form a multi-view training data synthesis pipeline, where 2D motion diffusion and multi-view optimization are used to generate consistent multi-view 2D pose sequences. The final stage trains a multi-view 2D motion diffusion model on these synthetic data to directly produce consistent multi-view 2D motions from a single 2D sequence input.
Together, these stages enable 3D motion reconstruction without requiring any 3D supervision.
While MVLift effectively reconstructs human motion from static, single-view videos, it cannot leverage Internet videos with dynamic cameras and limited viewpoint coverage during training. For human-object interaction (HOI) reconstruction, MVLift assumes access to precomputed 2D human and object keypoints reprojected from motion-capture data, without addressing reconstruction from real-world videos. These limitations motivate our new framework.

\myparagraph{Overview.} 
We propose \model, a unified framework that reconstructs both 3D human motion and human-object interactions (HOI) from monocular videos captured by dynamic cameras. 
An overview of \model is shown in \cref{fig:method}.
\model follows a two-stage pipeline:
(1) \emph{multi-view 2D synthetic data generation}, where we prepare training data with diverse camera trajectories (\cref{sec:training-data-preparation}); and
(2) \emph{multi-view 2D motion diffusion}, where we learn to generate consistent multi-view 2D motion from single-view 2D inputs using the synthesized data, and subsequently reconstruct world-coordinated 3D motion (\cref{sec:multiview-diffusion}, \cref{sec:hoi-reconstruction}).
For \textbf{human motion} such as gymnastics and martial arts that are rarely represented in existing motion-capture datasets, we leverage Internet videos to extract 2D keypoints and camera trajectories, and train a conditional single-view 2D diffusion model to synthesize multi-view training data.
For \textbf{human-object interactions (HOI)}, we follow the same two-stage pipeline but generate synthetic multi-view data by reprojecting existing 3D HOI motion-capture sequences~\cite{bhatnagar2022behave,li2023object,zhang2023neuraldome,huang2024intercap,zhao2024m}.

Built upon MVLift, our method introduces several key innovations. First, to accommodate dynamic cameras, we design a single-view 2D diffusion model conditioned on both camera trajectories and epipolar lines. Second, Internet videos of specific motion categories are typically recorded from limited forward-facing views, resulting in insufficient viewpoint coverage. We propose a hybrid training strategy that combines 2D motions extracted from videos and local 2D poses reprojected from reconstructed 3D motions obtained using off-the-shelf estimators. Third, we develop a multi-view 2D motion diffusion model conditioned on camera trajectories for human motion and extend it to HOI reconstruction from single-view 2D inputs. Finally, we further extend our HOI reconstruction approach to handle category‑specific object keypoints tracked from real‑world videos, enhancing its generalization at inference to diverse real‑world videos beyond models trained with object‑specific reprojected MoCap data in MVLift.

\subsection{Multi-View 2D Synthetic Data Generation}\label{sec:training-data-preparation}

\myfirstparagraph{Conditional Single-View 2D Motion Diffusion.} 
We begin by training a conditional single-view 2D motion diffusion model for each motion category. The conditioning terms include camera trajectories and epipolar lines. Camera trajectories provide awareness of global viewpoint motion over time, allowing the model to learn the 2D root translation under dynamic cameras. We represent the camera trajectories as a sequence of extrinsic parameters $\mathbf{C}=\{{\mathbf{C}_t}\}_{t=1}^T$, where each $\mathbf{C}_t\in\mathbb{R}^{4\times3}$ is normalized by removing the camera transformation of the initial frame. Epipolar lines encode pairwise geometric constraints between views, encouraging the model to learn cross-view consistency. Each epipolar line $\boldsymbol{l}=(a,b,c)^{\mathrm{T}}$ is defined by the 2D line equation $ax+by+c=0$. For every frame, we assign an epipolar line to each keypoint, passing through the keypoint and its corresponding epipole, resulting in a condition matrix $\mathbf{L}_t \in \mathbb{R}^{K \times 3}$. During training, we simulate several fixed epipoles based on sampled camera extrinsics, while at inference time the epipole is determined by the relative camera transformation between paired views.

Following the DDPM framework~\cite{ho2020denoising}, we adopt a forward diffusion process that progressively adds noise to clean 2D motions over $N$ steps:
\begin{equation}
    q(\mathbf{X}_n\vert\mathbf{X}_{n-1})=\mathcal{N}(\mathbf{X}_n;\sqrt{1-\beta_n}\mathbf{X}_{n-1},\beta_n\mathbf{I}),
\end{equation}
where $n\leq N$ denotes the diffusion step, and $\beta_n$ is the variance schedule controlling noise magnitude. The reverse denoising process is learned by a network $\mathbf{X}_\theta$, which learns to iteratively denoise samples across $N$ steps starting from $\mathbf{X}_N\sim\mathcal{N}(\mathbf{0},\mathbf{I})$ conditioned on the camera trajectories $\mathbf{C}$ and epipolar lines $\mathbf{L}$.  We reparameterize the prediction objective to directly estimate the clean sample $\mathbf{X}_0$. The network is optimized with an $L_1$ reconstruction loss:
\begin{equation}
    \mathcal{L}=\mathbb{E}_{\mathbf{X}_0,n}\big\Vert\mathbf{X}_0-\mathbf{X}_\theta(\mathbf{X}_n,n,\mathbf{C},\mathbf{L})\big\Vert_1.\label{eq:loss-diffusion}
\end{equation}
In line with prior works~\cite{tevet2023human}, we adopt a Transformer-based backbone~\cite{vaswani2017attention} for the denoising network $\mathbf{X}_\theta$. The conditioning inputs $\mathbf{C}$ and $\mathbf{L}$ are concatenated with the noisy keypoint sequence $\mathbf{X}_n$ along the feature dimension and embedded through an MLP encoder before feeding into the backbone.

Following \citet{li2025lifting}, we add line matching loss to encourage the 2D keypoints to align with their corresponding epipolar lines by minimizing the 2D point-line distance:
\begin{equation}
    \mathcal{L}_\text{line}=\sum_{t=1}^T\big\langle\mathbf{L}_t,(\hat{\mathbf{X}}_t,\mathbf{1})\big\rangle,\label{eq:loss-line}
\end{equation}
where $\hat{\mathbf{X}}_t$ denotes the 2D keypoints at frame $t$ after the denoising process.

\myparagraph{Hybrid Data Source Training.} 
Unlike standard datasets such as AIST++~\cite{li2021ai}, which provide multi-view videos with uniformly distributed camera viewpoints, Internet videos of specific motion categories like gymnastics are usually captured from a few forward-facing angles, resulting in limited viewpoint coverage. 

To mitigate this limitation, we introduce a hybrid training strategy that combines two complementary sources of 2D motion data:
(1) 2D keypoints extracted from real Internet videos, and
(2) 2D projections $\mathbf{X}^\text{proj}$ obtained by reprojecting reconstructed 3D motions from off-the-shelf estimators~\cite{shen2024world}.
Since these estimators are generally reliable only for local pose estimation, we use only their local 2D projections with the root joint aligned to the image center, while discarding global translation.

However, including $\mathbf{X}^\text{proj}$ in training biases the model toward learning motion patterns with limited global translation, since these sequences lack root movement. To address this, we decompose each 2D motion $\mathbf{X}$ into root translation $\mathbf{X}^\text{r}\in\mathbb{R}^{T\times2\times2}$---represented by the two hip joints---and local pose $\mathbf{X}^\text{l}\in\mathbb{R}^{T\times(K-2)\times2}$.
The global 2D motion $\mathbf{X}^\text{g}$ is then recovered by adding the average root translation (computed across the two hip joints) back to the local pose. With this representation, the diffusion loss is computed as in \cref{eq:loss-diffusion}, while the line‑matching loss is applied to the global 2D motion $\mathbf{X}^\text{g}$ defined in \cref{eq:loss-line}.

During training, $\mathbf{X}^\text{proj}$ is generated by projecting reconstructed 3D motions through randomly sampled camera viewpoints from the training set, augmented with a small set of predefined camera trajectories to increase viewpoint diversity. We compute the diffusion loss only for the local pose:
\begin{equation}
    \mathcal{L}^\text{proj}=\mathbb{E}_{\mathbf{X}_0,n}\big\Vert\mathbf{M}\odot\mathbf{X}_0-\mathbf{M}\odot\mathbf{X}_\theta(\mathbf{X}_n^\text{proj},n,\mathbf{C},\mathbf{L})\big\Vert_1,
\end{equation}
where $\mathbf{M}$ is a binary mask that excludes the two hip joints from the loss computation. The line matching loss is not applied to $\mathbf{X}^\text{proj}$.

\myparagraph{Multi-View 2D Motion Data Synthesis.} 
Leveraging the learned 2D motion prior, we employ score distillation sampling~\cite{poole2023dreamfusion} with multi-view consistency loss to prepare multi-view training data. Specifically, given a single-view sequence, we optimize $V-1$ additional 2D keypoint sequences from viewpoints evenly distributed along a circular ring around the input camera, resulting in a set of sequences $\{\mathbf{X}_v\}_{v=1}^{V}$. The gradient of the SDS loss, which encourages each $\mathbf{X}_v$ to conform to the learned diffusion prior, is computed as:
\begin{equation}
    \nabla_{\mathbf{X}_v}\mathcal{L}_\text{SDS}=\mathbb{E}_{n,\epsilon}\Big[w(n)\big(\epsilon_\theta(\mathbf{X}_{v,n},n,\mathbf{C},\mathbf{L})-\epsilon\big)\Big],
\end{equation}
where the weight $w(n)$ is determined by the noise level $n$. 

For two different views $u$ and $v$, we compute the epipolar lines $\mathbf{L}^{u\to v}$ in view $v$ using the 2D keypoint sequence $\mathbf{X}_u$ and the relative camera transformation between the two views. We then apply line matching loss to enforce that the 2D keypoints $\mathbf{X}_v$ satisfy the corresponding geometric constraints:
\begin{equation}
\mathcal{L}^{u\to v}_\text{line}=\sum_{t=1}^T\big\langle\mathbf{L}^{u\to v}_t,(\mathbf{X}^\text{g}_{v,t},\mathbf{1})\big\rangle,
\end{equation}
where $\mathbf{X}^\text{g}_{v,t}$ represents the recovered global motion from the decomposed representation. Unlike MVLift~\cite{li2025lifting}, we apply the line-matching loss only between adjacent views and between each view and the input view during score distillation sampling for computational efficiency.

After obtaining roughly consistent multi-view 2D pose sequences via SDS optimization, we recover 3D joint positions by minimizing multi-view reprojection errors.
The recovered 3D joints are then used to fit SMPL parameters~\cite{loper2015smpl} using VPoser~\cite{pavlakos2019expressive}, producing full-body 3D motion sequences.
Finally, we reproject the fitted 3D motions into four evenly distributed cameras to generate geometrically consistent multi-view 2D training data.

\subsection{Multi-View 2D Motion Diffusion}\label{sec:multiview-diffusion}

We train a multi-view 2D motion diffusion model using the data synthesized in \cref{sec:training-data-preparation} to generate multi-view 2D motion sequences from a single-view input.

\myparagraph{Data and Condition Representation.}
We train the multi-view diffusion model on global 2D keypoint sequences. 
Camera trajectories are represented in the same way as described in \cref{sec:training-data-preparation}. During inference, we extract 2D human keypoint sequences using ViTPose~\cite{xu2022vitpose} and estimate camera motion with MegaSaM~\cite{li2025megasam}. 

\myparagraph{Model Architecture.}
We extend the single-view 2D motion diffusion model introduced in \cref{sec:training-data-preparation}. The camera condition is embedded in the same way. The transformer backbone is further augmented with cross-view attention layers to enhance multi-view awareness following MVLift~\cite{li2025lifting}.

\subsection{HOI Motion Reconstruction}\label{sec:hoi-reconstruction}

For human-object interactions, we train the multi-view diffusion model on specific object categories (e.g., boxes and tables). The objects are represented by a set of manually designed 2D keypoints $\mathbf{O} \in \mathbb{R}^{T \times M \times 2}$, where $M$ denotes the number of keypoints. The corresponding 3D positions of these keypoints on the canonical object mesh are denoted as $\mathbf{P} = \{\mathbf{p}_i\}_{i=1}^M$. The object 2D keypoints $\mathbf{O}$ are concatenated with the human keypoints $\mathbf{X}$ to form a unified representation. During training, we randomly mask out a subset of $\mathbf{O}$ to handle partial occlusions and potential tracking failures.

\myparagraph{Inference on Real-World Videos.}
The human keypoints and camera motions are extracted in the same way as for human motion reconstruction, while the object keypoints are tracked using DELTA~\cite{ngo2025delta}. The 3D mesh of each object is captured using a free 3D scanner, providing high-fidelity geometry for subsequent reconstruction and alignment.

The SMPL parameters $\mathcal{H}=(\mathbf{r},\boldsymbol{\phi},\mathbf{\Theta})$ are obtained through the final optimization process described in \cref{sec:training-data-preparation}. For objects, we also start with obtaining the 3D object keypoints $\mathbf{Q}\in\mathbb{R}^{T\times M\times3}$ by minimizing the multi-view reprojection error. Using the reconstructed $\mathbf{Q}$ and the predefined canonical keypoints $\mathbf{P}$ on the object mesh, we then estimate the object pose $\mathcal{O}_t = \{ \mathbf{r}_t, \mathbf{t}_t, s \}$, where $\mathbf{r}_t \in \mathbb{R}^6$ is the 6D rotation~\cite{zhou2019continuity}, $\mathbf{t}_t \in \mathbb{R}^3$ is the translation, and $s \in \mathbb{R}^+$ is a global scale factor, as detailed in \cref{sec:object-motion-optimization}.

\section{Experiments}

We evaluate \model on human motion reconstruction and human-object interaction reconstruction. Training and evaluation are conducted using both MoCap datasets with 3D ground truth and our collected video-only datasets.

\subsection{Datasets and Evaluation Metrics} 

\myfirstparagraph{Human Motion Reconstruction.}
For human motion lifting, we evaluate our method on the \textbf{AIST++}~\cite{li2021ai} dataset and two newly collected video-only datasets of \textbf{gymnastics} and \textbf{martial arts}. We train and test \model on these datasets separately. AIST++ provides multi-view RGB videos with corresponding 3D dance motions. We conduct evaluations under two settings: (1) following MVLift~\cite{li2025lifting}, we use its processed version of the AIST++ dataset for training and testing, where only one camera view is available for each sequence; and (2) since AIST++ does not include dynamic camera videos, we synthesize additional training and evaluation data by projecting the 3D ground truths with simulated moving cameras to assess \model's capability under dynamic cameras.

The collected gymnastics and martial arts videos feature highly dynamic motions involving complex body movements, typically captured with moving cameras and limited view coverage. Such sequences are rarely present in existing MoCap datasets. Evaluating these challenging cases allows us to assess \model’s ability to handle scenarios that are difficult for methods relying on MoCap data. We process the collected videos using ViTPose~\cite{xu2022vitpose} to obtain 2D pose sequences and estimate camera poses with MegaSaM~\cite{li2025megasam}.

We follow the evaluation protocol of MVLift~\cite{li2025lifting}. We project the reconstructed 3D motion to 2D and calculate the 2D joint position error ($\textbf{J}_\text{2D}$) and the 2D joint position error with the 2D root joint translated to the image center ($\textbf{J}^\text{C}_\text{2D}$). To assess motion realism, we train a 2D motion feature extractor for each dataset and compute the Fréchet Inception Distance ($\textbf{FID}$) on 2D motion features. For datasets with 3D ground truth, we further evaluate performance using the root translation error ($\textbf{T}_{\text{root}}$), the mean per-joint position error (\textbf{MPJPE}), and the Procrustes-aligned mean per-joint position error (\textbf{PA-MPJPE}). In addition, we assess the foot sliding score ($\textbf{FS}$) following \citet{he2022nemf}.

\myparagraph{HOI Motion Reconstruction.}
We evaluate \model’s capability for HOI reconstruction on the \textbf{BEHAVE} dataset~\cite{bhatnagar2022behave}. We train separate models for three object categories: \textit{box}, \textit{chair}, and \textit{table}. For training data, we use interaction sequences involving these objects from the BEHAVE training set, together with additional data from InterCap~\cite{huang2024intercap}, HODome~\cite{zhang2023neuraldome}, IMHD~\cite{zhao2024m}, and OMOMO~\cite{li2023object}. Following the evaluation protocol used for AIST++, we perform experiments under two settings: (1) static‑camera setting using the original camera poses from the BEHAVE test set, and (2) synthetic dynamic‑camera setting obtained via simulated camera motion. Object keypoint sequences are obtained by reprojecting 3D HOIs from the BEHAVE dataset. We also provide an alternative optimization-based method for extracting object 2D keypoints (\cref{sec:hoi-video-processing}), with qualitative results available on our project page.

Following the human motion lifting evaluation, we assess human motion accuracy using $\textbf{T}_{\text{root}}$, \textbf{MPJPE}, and \textbf{PA-MPJPE}. For object motions, we evaluate the object translation error ($\textbf{T}^\text{O}_{\text{root}}$) and the object mean per-joint position error (\textbf{O-MPJPE}), which respectively quantify the translation and keypoint accuracy of reconstructed object movements.

\subsection{Human Motion Reconstruction}

\begin{table}[t!]
    \centering
    \small
    \setlength{\tabcolsep}{3pt}
    \resizebox{\linewidth}{!}{%
        \begin{tabular}{lccccccc}
            \toprule
            Method & $\textbf{J}_\text{2D}$ & $\textbf{J}^\text{C}_\text{2D}$ & $\textbf{FID}$ & $\textbf{T}_\text{root}$  & $\textbf{MPJPE}$  & $\textbf{PA-MPJPE}$ & $\textbf{FS}$ \\
            \midrule
            SMPLify~\cite{bogo2016keep} & 24.3 & 14.7 & 2.4 & 90.3 & 174.7 & 132.8 & 1.548 \\
            WHAM~\cite{shin2024wham} & 75.5 & 22.1 & 3.1 & 164.3 & \underline{104.8} & \underline{75.1} & 0.579 \\
            GVHMR~\cite{shen2024world} & 106.4 & 20.3 & 2.9 & 143.0 & \textbf{97.6} & \textbf{64.4} & 0.547 \\
            MVLift~\cite{li2025lifting} & \underline{17.5} & \underline{14.3} & \underline{2.2} & \underline{67.6} & 110.7 & 79.2 & \textbf{0.471} \\
            \model (ours) & \textbf{16.6} & \textbf{13.3} & \textbf{2.1} & \textbf{64.9} & 108.0 & 82.3 & \underline{0.475} \\
            \midrule
            SMPLify~\cite{bogo2016keep} & 24.7 & \underline{14.9} & 2.7 & 90.9 & 175.2 & 134.6 & 1.530 \\
            MVLift~\cite{li2025lifting} & \underline{18.0} & \underline{14.9} & \underline{2.1} & \underline{64.9} & \underline{122.1} & \underline{94.3} & \underline{0.487} \\
            \model (ours) & \textbf{16.7} & \textbf{13.7} & \textbf{2.0} & \textbf{64.2} & \textbf{109.3} & \textbf{83.0} & \textbf{0.446} \\
            \bottomrule
        \end{tabular}
    }
    \vspace{-0.25cm}
    \caption{\textbf{Quantitative evaluation on the AIST++ dataset~\cite{li2021ai}} under (1) static‑camera setup (upper) and (2) dynamic‑camera setup (lower). \model achieves competitive 3D joint accuracy and improved root translation estimation while maintaining robustness under dynamic camera.}
    \label{tab:aist}
    \vspace{-2mm}
\end{table}

\begin{table}[t!]
    \centering
    \small
    \setlength{\tabcolsep}{4pt}
    \resizebox{\linewidth}{!}{%
        \begin{tabular}{lcccccccc}
            \toprule
            \multirow{2}{*}{Method} & \multicolumn{4}{c}{\textbf{Gymnastics}} & \multicolumn{4}{c}{\textbf{Martial Arts}} \\
            \cmidrule(r){2-5} \cmidrule(l){6-9} & $\textbf{J}_\text{2D}$ & $\textbf{J}^\text{C}_\text{2D}$ & $\textbf{FID}$ & $\textbf{FS}$ & $\textbf{J}_\text{2D}$ & $\textbf{J}^\text{C}_\text{2D}$ & $\textbf{FID}$ & $\textbf{FS}$ \\
            \midrule
            SMPLify~\cite{bogo2016keep} & 39.2 & \underline{16.0} & \underline{11.2} & 0.463 & 48.0 & 13.2 & 5.8 & 0.397 \\
            WHAM~\cite{shin2024wham} & 88.6 & 21.7 & 16.4 & 0.245 & 87.4 & 22.3 & 10.7 & 0.212 \\
            GVHMR~\cite{shen2024world} & 71.5 & 18.8 & 13.0 & \textbf{0.061} & 66.3 & 15.9 & 6.0 & \textbf{0.056} \\
            MVLift~\cite{li2025lifting} & \underline{33.1} & 17.0 & \underline{11.2} & 0.188 & \underline{24.6} & \underline{12.0} & \underline{4.6} & 0.145 \\
            \model (ours) & \textbf{21.6} & \textbf{11.4} & \textbf{10.9} & \underline{0.152} & \textbf{15.1} & \textbf{9.8} & \textbf{3.6} & \underline{0.136} \\
            \bottomrule
        \end{tabular}
    }
    \vspace{-0.25cm}
    \caption{\textbf{Quantitative evaluation on our collected Internet videos.} \model outperforms all baselines across most metrics, demonstrating the plausibility of our method on Internet videos.}
    \label{tab:internet-video}
    \vspace{-2mm}
\end{table}

\begin{table}[t!]
    \centering
    \small
    \setlength{\tabcolsep}{8pt}
    \resizebox{\linewidth}{!}{%
        \begin{tabular}{lcc}
            \toprule
            Method & \textbf{Ground Contact }& \textbf{Motion Quality} \\
            \midrule
            vs. SMPLify~\cite{bogo2016keep} & 84.2\% & 85.0\% \\
            vs. WHAM~\cite{shin2024wham} & 75.6\% & 74.2\% \\
            vs. GVHMR~\cite{shen2024world} & 75.5\% & 61.7\% \\
            vs. MVLift~\cite{li2025lifting} & 61.3\% & 65.4\% \\
            vs. \model w/o Hybrid & 65.6\% & 66.3\% \\
            \bottomrule
        \end{tabular}
    }
    \vspace{-0.25cm}
    \caption{\textbf{Human study on reconstructed human motions from our collected Internet videos.} Participants prefer our reconstruction results for their better ground contact and motion quality.}
    \label{tab:human-study}
    \vspace{-4mm}
\end{table}

\begin{figure*}[t!]
    \centering
    \includegraphics[width=1\textwidth]{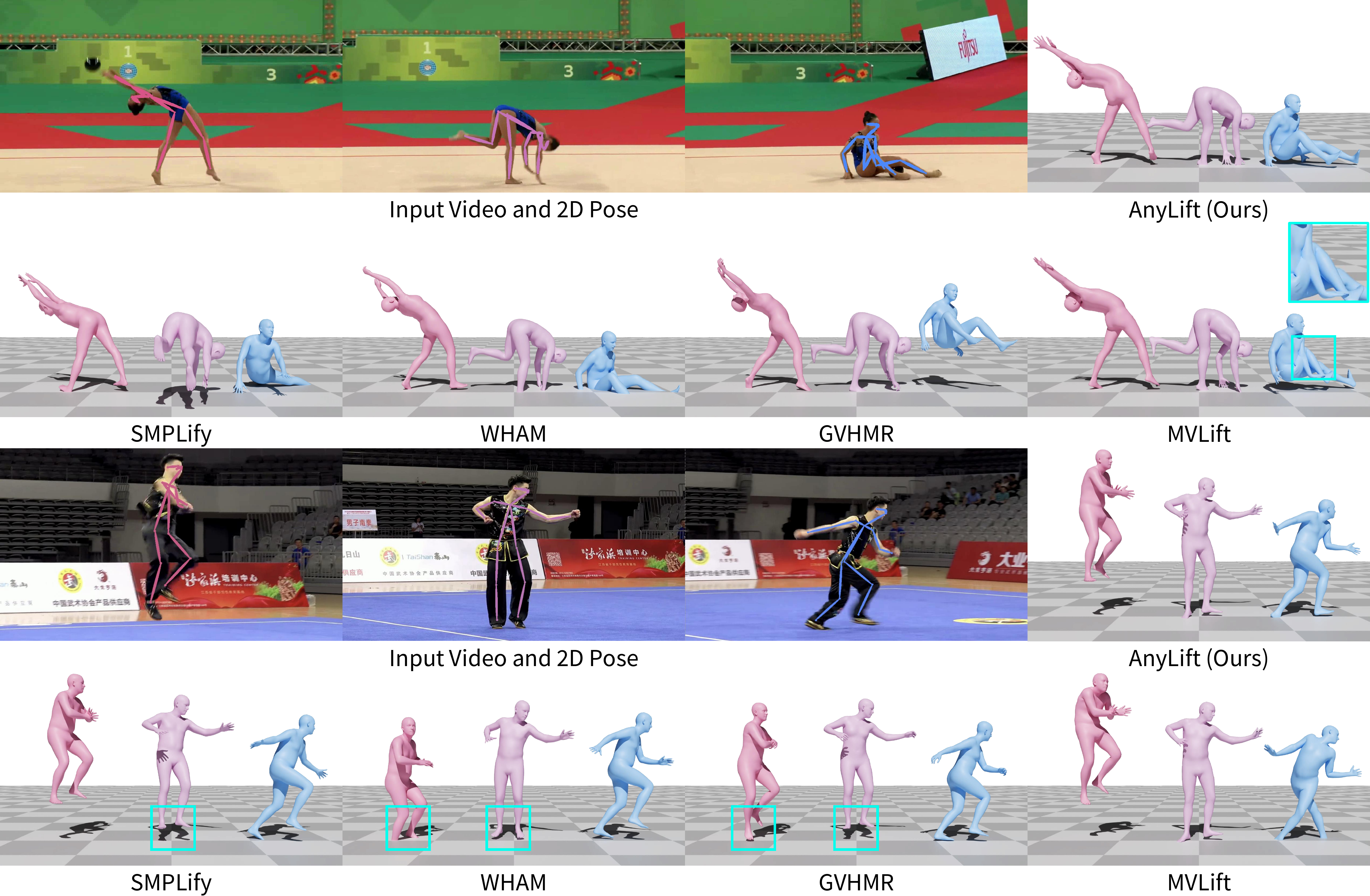}
    \vspace{-0.6cm}
    \caption{\textbf{Qualitative comparison of human motion reconstruction on our collected Internet videos.} \model produces more plausible motions, mitigating the root trajectory errors, inaccurate local body pose, and self‑penetration artifacts observed in baselines.}
    \label{fig:res-human}
\end{figure*}

\begin{figure*}[t!]
    \centering
    \includegraphics[width=1\textwidth]{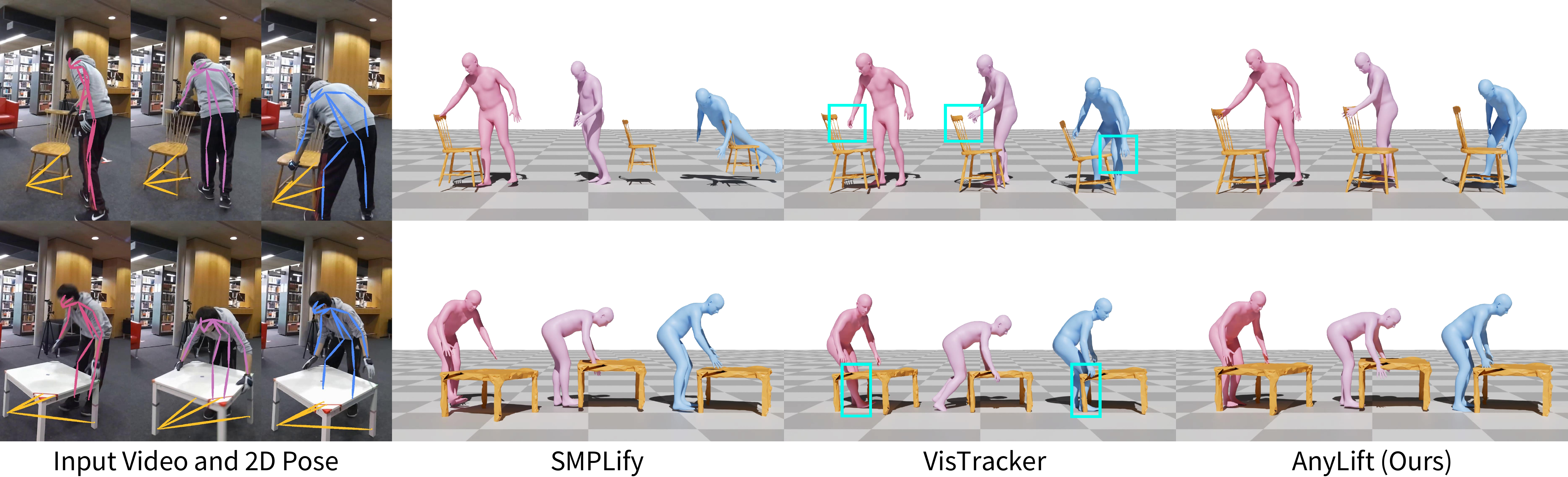}
    \vspace{-0.6cm}
    \caption{\textbf{Qualitative comparison of HOI reconstruction on the BEHAVE~\cite{bhatnagar2022behave} dataset.} We show results on two object categories, \textit{chair} and \textit{table}. \model produces coherent and physically plausible human-object interactions with accurate contact and minimal penetration.}
    \label{fig:res-hoi}
    \vspace{-2mm}
\end{figure*}

\myfirstparagraph{Baselines.}
We compare our method with two categories of baselines: methods that do not rely on 3D motion data during training, and methods trained with 3D motion ground truth. SMPLify~\cite{bogo2016keep} estimates SMPL~\cite{loper2015smpl} parameters by optimizing 2D reprojection objectives without requiring any training. In contrast, WHAM~\cite{shin2024wham} and GVHMR~\cite{shen2024world} leverage AMASS~\cite{mahmood2019amass} together with several other datasets for training. We further adapt MVLift~\cite{li2025lifting} to the dynamic camera setting by first running its original pipeline and then applying the estimated camera motion to its outputs.

\myparagraph{Results.}
We report quantitative results on the AIST++ dataset~\cite{li2021ai} in \cref{tab:aist}. Under the original MVLift evaluation setting with static cameras (upper), \model outperforms all baselines across most metrics, including the original MVLift implementation. In terms of 3D joint position errors, our method achieves comparable accuracy to WHAM and GVHMR, which require training on AMASS, while showing a substantial improvement in root translation accuracy. Under the synthetic dynamic-camera setting (lower), \model significantly outperforms both SMPLify and MVLift across all metrics. Notably, \model maintains similar accuracy on 3D joint position errors between the two settings, highlighting its robustness to real-world camera motion. The quantitative results on our collected Internet videos are presented in \cref{tab:internet-video}. \model achieves superior performance over all baseline methods across most metrics.

\begin{table*}[t!]
    \centering
    \small
    \setlength{\tabcolsep}{2pt}
    \resizebox{\linewidth}{!}{%
        \begin{tabular}{lccccccccccccccc}
            \toprule
            \multirow{2}{*}{Method} & \multicolumn{5}{c}{\textbf{Box}} & \multicolumn{5}{c}{\textbf{Chair}} & \multicolumn{5}{c}{\textbf{Table}} \\ 
            \cmidrule(r){2-6} \cmidrule(l){7-11} \cmidrule(l){12-16} & $\textbf{T}_\text{root}$  & $\textbf{MPJPE}$  & $\textbf{PA-MPJPE}$ & $\textbf{T}^\text{O}_{\text{root}}$ & $\textbf{O-MPJPE}$ & $\textbf{T}_\text{root}$  & $\textbf{MPJPE}$  & $\textbf{PA-MPJPE}$ & $\textbf{T}^\text{O}_{\text{root}}$ & $\textbf{O-MPJPE}$ & $\textbf{T}_\text{root}$  & $\textbf{MPJPE}$  & $\textbf{PA-MPJPE}$ & $\textbf{T}^\text{O}_{\text{root}}$ & $\textbf{O-MPJPE}$ \\
            \midrule
            SMPLify~\cite{bogo2016keep} & 78.12 & 114.56 & 86.64 & 795.06 & \underline{207.77} & 64.86 & 102.76 & 80.75 & 438.98 & 145.35 & 70.06 & 109.31 & 86.10 & 385.60 & \underline{124.86} \\
            VisTracker~\cite{xie2023visibility} & \underline{51.72} & \underline{54.40} & \underline{50.40} & \underline{143.59} & 359.50 & \underline{52.37} & \underline{72.72} & \underline{67.84 } & \underline{90.18 } & \underline{134.95} & \underline{65.18} & \underline{85.51} & \underline{82.73} & \underline{177.17} & 540.96 \\
            \model (ours) & \textbf{24.61} & \textbf{42.68} & \textbf{35.62} & \textbf{95.38} & \textbf{32.98} & \textbf{22.48} & \textbf{53.58} & \textbf{43.85} & \textbf{75.93} & \textbf{34.54} & \textbf{26.05} & \textbf{48.34} & \textbf{39.06} & \textbf{77.53} & \textbf{51.28} \\
            \midrule
            SMPLify~\cite{bogo2016keep} & 82.21 & 126.12 & 89.85 & 669.64 & 185.48 & 63.37 & 107.22 & 81.73 & 555.70 & 165.90 & 77.19 & 119.42 & 78.03 & 514.62 & 149.91 \\
            \model (ours) & \textbf{29.99} & \textbf{43.60} & \textbf{39.02} & \textbf{96.76} & \textbf{33.96} & \textbf{23.87} & \textbf{57.82} & \textbf{45.25} & \textbf{79.16} & \textbf{44.60} & \textbf{28.09} & \textbf{54.60} & \textbf{42.91} & \textbf{88.93} & \textbf{56.97} \\
            \bottomrule
        \end{tabular}
    }
    \vspace{-0.25cm}
    \caption{\textbf{Quantitative evaluation on the BEHAVE dataset~\cite{bhatnagar2022behave}} under (1) static‑camera setup (upper) and (2) dynamic‑camera setup (lower). \model outperforms all baselines across object categories and achieves robust performance under dynamic‑camera conditions.}
    \label{tab:behave_static_dynamic}
    \vspace{-4mm}
\end{table*}

We show qualitative comparisons in \cref{fig:res-human}. SMPLify produces 3D poses with abrupt and unrealistic changes due to depth ambiguity arising from optimization based solely on 2D joint positions. Although WHAM and GVHMR predict plausible local poses, they often yield implausible root trajectories, leading to noticeable penetration with the ground plane, as shown in the two examples. MVLift suffers from self‑penetration (first example) and yields severely distorted leg poses in the second example. In contrast, \model reconstructs stable and plausible human motions with accurate global trajectories and consistent body poses across frames.

\myparagraph{Human Perceptual Study.}
We conduct a human perceptual study using the two‑alternative forced choice (2AFC) method. A total of 300 participants are recruited via the Prolific platform. In each trial, participants are presented with reconstruction results from different methods and asked to choose the one that exhibits better ground contact and motion quality. As shown in \cref{tab:human-study}, participants consistently prefer our results over the baselines, aligning well with the quantitative metrics and qualitative comparisons.

\subsection{HOI Motion Reconstruction}

\myfirstparagraph{Baselines.}
We compare against two representative baselines. 
SMPLify~\cite{bogo2016keep} reconstructs 3D human and object motions by optimizing SMPL parameters and object poses to match 2D keypoints without any training. 
VisTracker~\cite{xie2023visibility} jointly tracks 3D humans, objects, and contacts from monocular videos using a visibility-aware object pose network. 

\myparagraph{Results.}  
Quantitative results on the BEHAVE dataset are shown in \cref{tab:behave_static_dynamic}. Across all object categories (\textit{box}, \textit{chair}, \textit{table}) and both static and dynamic camera settings, \model consistently outperforms the baselines in all metrics. 
SMPLify, which relies purely on optimization based on 2D reprojection objectives, struggles to recover accurate 3D motion and translation, often converging to suboptimal local minima when 2D evidence is ambiguous. 
VisTracker suffers from temporal jitter and ambiguity when handling symmetric objects, as its visibility-aware module may yield inconsistent pose predictions across frames. 
These issues lead to large errors in object-related metrics (\(\text{T}^\text{O}_\text{root}\) and \(\text{O-MPJPE}\)). 
In contrast, \model jointly predicts human and object dynamics within a unified framework, achieving stable and accurate reconstruction even under challenging dynamic-camera conditions. 
The small performance gap between static and dynamic settings further demonstrates its strong generalization to realistic camera motion.

Qualitative comparisons on the BEHAVE dataset are presented in \cref{fig:res-hoi}. We show two object categories: \textit{chair} and \textit{table}. Similar to human motion reconstruction, SMPLify fails to produce reasonable human-object interaction motions due to depth ambiguity from relying solely on 2D joints and incorrectly models the relative transformation between the human and the object. VisTracker also struggles to generate plausible interactions; in the chair example, even with minimal or no occlusion, it fails to capture correct hand-chair contact, while in the table example, it exhibits severe object penetration. In contrast, our approach produces coherent and plausible human-object interactions with accurate contact and minimal penetration under both categories. We further present qualitative results on real-world videos captured in natural environments in \cref{fig:teaser}.

\begin{table}[t!]
    \centering
    \small
    \setlength{\tabcolsep}{6pt}
    \resizebox{\linewidth}{!}{%
        \begin{tabular}{lcccccc}
            \toprule
            \multirow{2}{*}{Method} & \multicolumn{3}{c}{\textbf{Gymnastics}} & \multicolumn{3}{c}{\textbf{Martial Arts}} \\
            \cmidrule(r){2-4} \cmidrule(l){5-7} & $\textbf{J}_\text{2D}$ & $\textbf{J}^\text{C}_\text{2D}$ & $\textbf{FID}$ & $\textbf{J}_\text{2D}$ & $\textbf{J}^\text{C}_\text{2D}$ & $\textbf{FID}$ \\
            \midrule
            \model w/o Hybrid & 36.1 & 18.7 & 11.2 & 25.2 & 12.7 & 4.3 \\
            \model (ours) & \textbf{21.6} & \textbf{11.4} & \textbf{10.9} & \textbf{15.1} & \textbf{9.8} & \textbf{3.6} \\
            \bottomrule
        \end{tabular}
    }
    \vspace{-0.25cm}
    \caption{\textbf{Ablation study on our collected Internet videos.} Performance drops across all metrics without incorporating local 2D poses from diverse viewpoints.}
    \label{tab:ablation}
    \vspace{-4mm}
\end{table}

\subsection{Ablation Study}

We conduct an ablation study on our collected Internet video datasets, which are captured with limited camera views. We remove the hybrid training strategy that incorporates local poses estimated by GVHMR and train \model solely on 2D motion sequences extracted from the videos. The quantitative results are presented in \cref{tab:ablation}. Our full model achieves superior performance across all metrics compared to the ablated variants. Together with the comparison to GVHMR~\cite{shen2024world} in \cref{tab:internet-video}, these results demonstrate that combining estimated local 2D poses from diverse viewpoints with global 2D motions directly extracted from videos is a more effective design choice.

\section{Conclusion}
We presented \model, a unified framework for reconstructing world-grounded human motion and human-object interactions (HOI) from Internet and in-the-wild videos with dynamic cameras. We addressed the problem using a two-stage framework that first synthesizes multi-view 2D motion data and then trains a camera-conditioned multi-view diffusion model on the generated data to reconstruct globally consistent 3D motion and interactions in the world coordinate frame. We demonstrated the effectiveness of our approach on newly collected Internet videos featuring complex human motions, as well as on our captured in-the-wild HOI videos.

\pagebreak

\clearpage

\myparagraph{Acknowledgement.} We thank Chen Geng, Yanzhe Lyu, Yichao Zhou, and Yongqian Peng for fruitful discussions. We also extend our sincere thanks to Yazhou Zhang, Cyrus Zhou, and Di Fan for their support in data collection. This work is in part supported by the Stanford Institute for Human-Centered AI (HAI) and ONR MURI N00014-22-1-2740.

{
    \small
    \bibliographystyle{ieeenat_fullname}
    \bibliography{main}
}

\clearpage
\appendix
\renewcommand\thefigure{S\arabic{figure}}
\setcounter{figure}{0}
\renewcommand\thetable{S\arabic{table}}
\setcounter{table}{0}
\renewcommand\theequation{S\arabic{equation}}
\setcounter{equation}{0}
\pagenumbering{arabic}%
\renewcommand*{\thepage}{S\arabic{page}}
\setcounter{footnote}{0}
\setcounter{page}{1}

\section{Overview}

In this supplementary material, we provide additional details on implementation (\cref{sec:implementation-details}) and video data processing (\cref{sec:data-processing-details}). We highly recommend viewing our \href{https://awfuact.github.io/anylift/}{\textit{project page}} for compelling demonstrations.

\section{Additional Implementation Details}\label{sec:implementation-details}

\subsection{Details on Model Training}

We train the single-view 2D motion diffusion model for 300,000 steps and the multi-view diffusion model for 120,000 steps. The Adam optimizer\cite{diederik2015adam} is used with a learning rate of $1\times10^{-4}$, a batch size of 64, and no weight decay. All training is performed on a single NVIDIA L40S GPU. Model checkpoints are saved every 20,000 steps, and the final model is selected based on validation performance. During the training process of multi-view diffusion, we randomly mask out a subset of input keypoints to handle partial occlusions and potential tracking failures.

For the hybrid data-source training, we use a data ratio of 2:1 between 2D keypoints extracted from real Internet videos and local 2D projections reprojected from reconstructed 3D motions.

\subsection{Details on Object Motion Optimization}\label{sec:object-motion-optimization}

We provide details on how we recover the 3D object motion from our sampled multi-view 2D results. We start with obtaining the 3D object keypoints $\mathbf{Q}\in\mathbb{R}^{T\times M\times3}$ by minimizing the multi-view reprojection error. Using the reconstructed $\mathbf{Q}$ and the predefined canonical keypoints $\mathbf{P}\in\mathbb{R}^{M\times3}$ on the object mesh, we then estimate the object pose $\mathcal{O}_t=\{\mathbf{r}_t,\mathbf{t}_t,s\}$, where $\mathbf{r}_t\in\mathbb{R}^6$ is the 6D rotation~\cite{zhou2019continuity}, $\mathbf{t}_t\in\mathbb{R}^3$ is the translation, and $s \in \mathbb{R}^+$ is a global scale factor. We denote the mapping from $\mathbf{r}_t$ to the rotation matrix as $\mathbf{R}_t=\text{Rot}(\mathbf{r}_t)\in\mathrm{SO}(3)$.

We first initialize each frame independently by aligning the canonical keypoints $\mathbf{P}$ with the observed 3D keypoints $\mathbf{Q}_t$ through a rigid transformation and scale estimation as:
\begin{equation}
\hat{\mathbf{r}}_t,\hat{\mathbf{t}}_t=\arg\min_{\mathbf{r}_t,\mathbf{t}_t}\sum_{i=1}^{M}\Big\Vert\,\hat{s}\,\text{Rot}(\mathbf{r}_t)\mathbf{p}_i+\mathbf{t}_t-\mathbf{q}_{i,t}\Big\Vert_2,
\end{equation}
where the global scale is computed from the distance ratio of a reference keypoint pair in the observed and canonical spaces:
\begin{equation}
\hat{s}=\frac{\Vert\,\mathbf{q}_{1,t}-\mathbf{q}_{2,t}\Vert_2}{\Vert\,\mathbf{p}_1-\mathbf{p}_2\Vert_2}.
\end{equation}

This initialization yields per-frame pose estimates without temporal coupling. After initialization, we further optimize the object pose sequence with the object fitting loss:
\begin{equation}
\mathcal{L}_{\text{fit}}^\text{obj}=\frac{1}{TM}\sum_{t=1}^{T}\sum_{i=1}^{M}m_i\Big\Vert\,s\,\text{Rot}(\mathbf{r}_t)\mathbf{p}_i+\mathbf{t}_t-\mathbf{q}_{i,t}\Big\Vert_2,
\end{equation}
where $m_i\in\{0,1\}$ is a point-wise visibility mask shared across all frames.

Additionally, we apply a regularization loss on consecutive rotations to ensure temporal smoothness:
\begin{equation}
\mathcal{L}_{\text{smooth}}^\text{obj}=\frac{1}{T-1}\sum_{t=1}^{T-1}\Big\Vert\,\mathbf{r}_t-\mathbf{r}_{t+1}\Big\Vert_2.
\end{equation}

The optimization loss for object motion is defined as:
\begin{equation}
\mathcal{L}_{\text{obj}}=\mathcal{L}_{\text{fit}}^\text{obj}+\lambda_{\text{smooth}}^\text{obj}\mathcal{L}_{\text{smooth}}^\text{obj}.
\end{equation}

The Adam optimizer~\cite{diederik2015adam} is employed for all optimization stages. To reconstruct the 3D keypoint sequence from the sampled multi-view 2D motions, we run 500 optimization iterations with a learning rate of $0.01$. During the object pose fitting stage, we perform 2,000 iterations with a learning rate of $0.05$.

\subsection{Details on Predefined Camera Trajectory}\label{sec:predefined-camera}

To enrich camera trajectory diversity, we predefine six camera movement modes: zoom in, zoom out, move left, move right, rotate clockwise, and rotate counterclockwise. Each mode generates a sequence of camera transformations simulating the corresponding motion pattern. For each sequence, we randomly determine whether the camera keeps moving in one direction throughout the sequence or moves back to its original position after reaching the maximum displacement. In addition, with a certain probability, the camera tracks the pelvic joint to keep the human roughly centered in the view.

During training, the camera configuration is dynamically determined at each step. For each 3D motion sequence, we randomly choose to use either a camera trajectory extracted from real Internet videos (70\%) or one of the predefined camera trajectories (30\%) to reproject it into 2D for training.

\subsection{Details on Experimental Setup}

We provide details on how we synthesize additional training and evaluation data by projecting the 3D ground-truth motions using simulated moving cameras on the AIST++~\cite{li2021ai} and BEHAVE~\cite{bhatnagar2022behave} datasets. Following our hybrid data-source training, we reproject the 3D data using camera viewpoints sampled from a large camera motion base estimated from our gymnastic and martial arts videos, and we also incorporate a small set of predefined camera trajectories (\cref{sec:predefined-camera}) with a ratio of 7:3 between the two sources.

We ensure that camera motions used for training do not overlap with those in the testing set. The estimated camera motion base is inherently divided into separate training and testing subsets to enforce this separation. For the predefined cameras, the range of motion, the decision of whether to return to the origin, and whether to track the human pelvic joint are all randomized, ensuring that no camera trajectory in the testing phase exactly matches any trajectory seen during training.

\begin{figure}[t!]
    \centering
    \begin{subfigure}{\linewidth}
        \centering
        \includegraphics[width=\linewidth]{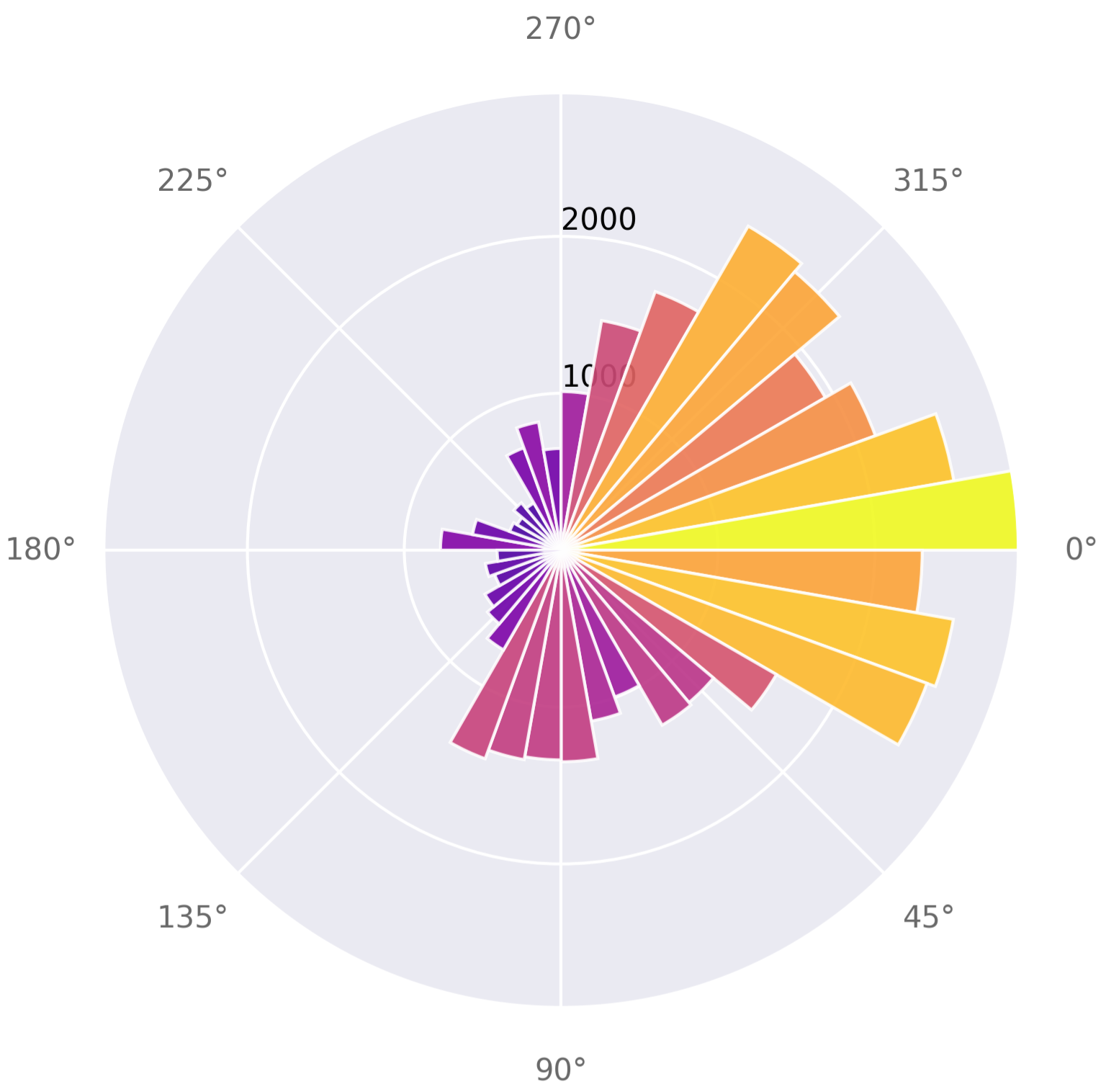}
    \end{subfigure}
    \\
    \begin{subfigure}{\linewidth}
        \centering
        \includegraphics[width=\linewidth]{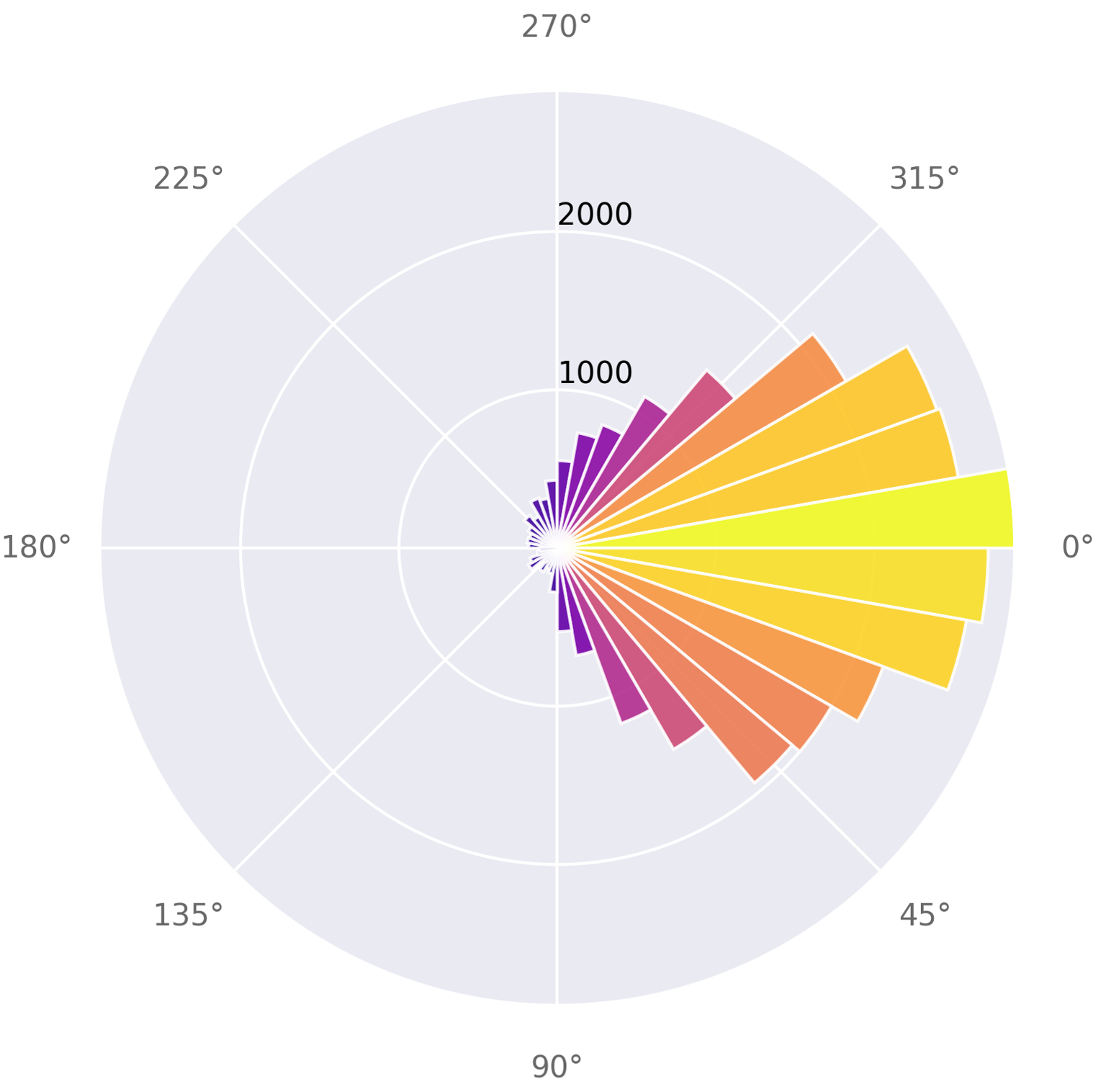}
    \end{subfigure}
    \caption{\textbf{Facing direction distributions of estimated humans} in the gymnastics (upper) and martial arts (lower) videos under the camera coordinate system. The angular axis indicates the facing direction and the radial axis represents number of frames.}
    \label{fig:direction}
\end{figure}

\section{Additional Details on Data Processing}\label{sec:data-processing-details}

\subsection{Details on Internet Video Processing}

We provide details on how we process the manually collected Internet videos. We first filter out low-quality videos, such as those containing advertisements or long idle segments. The remaining videos are then divided into 10-second clips, resulting in roughly 4,000 gymnastics clips and 5,000 martial arts clips. In subsequent processing, we further filter out these clips to handle failures in human tracking and 2D pose estimation. Segments with low pose confidence score or severe 2D jittering are discarded. After the filtering process, we obtain around 1,600 gymnastics clips and 3,000 martial arts clips that are used for training and evaluation. The processed data are then split into training and testing sets with a 9:1 ratio.

As shown in \cref{fig:direction}, we plot the facing directions of the estimated humans in the gymnastics and martial arts videos in the camera coordinate system using polar plots, where the angular axis represents the facing direction and the radial axis corresponds to the number of frames. The results show that the facing directions in both datasets are mostly concentrated within the same semicircle, which supports our observation that the viewpoint coverage of online videos is severely limited.

\subsection{Details on HOI Video Processing}\label{sec:hoi-video-processing}

\myfirstparagraph{Processing Videos from the BEHAVE Dataset.}
The BEHAVE dataset~\cite{bhatnagar2022behave} involves complex object motions, particularly frequent rotations, which makes off-the-shelf tracking methods~\cite{ngo2025delta} unreliable. To obtain more reliable 2D keypoints, we avoid tracking in the image plane and instead recover the per-frame object pose by aligning the object mesh to the observed object mask using a point sampling, projection-based method.

Concretely, for each object, we start from its template mesh and pre-select a small set of mesh vertices as semantic keypoints. In each frame, we first obtain an object segmentation mask using Grounded-SAM~\cite{ren2024grounded}, and randomly sample 5,000 2D points $\mathcal{Q}=\{\mathbf{q}_j\}$ from the foreground pixels. We also randomly sample 5,000 3D surface points $\mathcal{X}=\{\mathbf{x}_i\}$ from the object mesh. For projection, we fix a pinhole camera model with focal length $f_x=f_y=1000$ and principal point $(c_x,c_y) = (\tfrac{W}{2}, \tfrac{H}{2})$ determined by the image width $W$ and height $H$. With these intrinsics and an unknown $\mathrm{SE}(3)$ pose $(\mathbf{R},\mathbf{t})$ of the object in the camera coordinate system, each 3D point is projected to the image as
\begin{equation}
\mathbf{p}_i = \Pi(\mathbf{R} \mathbf{x}_i + \mathbf{t}),
\end{equation}
where $\Pi(\cdot)$ denotes the pinhole projection. Let $\mathcal{P}=\{\mathbf{p}_i\}$ be the set of projected points. We estimate $(\mathbf{R},\mathbf{t})$ by minimizing a symmetric 2D Chamfer distance between $\mathcal{P}$ and $\mathcal{Q}$:
\begin{align}
\mathcal{L}_\text{chamfer}&= \frac{1}{|\mathcal{P}|} \sum_{\mathbf{p}_i \in \mathcal{P}} \min_{\mathbf{q}_j \in \mathcal{Q}} \|\mathbf{p}_i - \mathbf{q}_j\|_2^2\nonumber\\
&+ \frac{1}{|\mathcal{Q}|} \sum_{\mathbf{q}_j \in \mathcal{Q}} \min_{\mathbf{p}_i \in \mathcal{P}} \|\mathbf{q}_j - \mathbf{p}_i\|_2^2 .
\end{align}

In implementation, we parameterize $\mathbf{R}$ using a continuous 6D rotation representation~\cite{zhou2019continuity} and initialize $\mathbf{t}$ heuristically from the 2D mask bounding box and the object’s 3D extent, which stabilizes optimization under large rotations. For the first frame of each sequence, we perform 200 random restarts of $\mathbf{R}$ and retain the solution with the lowest Chamfer loss. For each subsequent frame, we use the optimized pose from the previous frame as initialization, allowing the optimizer to refine the pose smoothly over time. After convergence, we apply the estimated $(\mathbf{R}, \mathbf{t})$ to the predefined semantic keypoints and project them into the image, producing temporally consistent 2D object keypoints on BEHAVE RGB videos. The Adam optimizer~\cite{diederik2015adam} is adopted in this optimization process.

\myparagraph{Processing Captured Real-World Videos.}
For the videos we captured, we manually select a small set of visible object keypoints in the first RGB frame and track them across the sequence using DELTA~\cite{ngo2025delta}. The reconstructed human-object interactions and tracked object keypoints are visualized on our project webpage.

\end{document}